\documentclass{article}

\usepackage{arxiv}

\usepackage[utf8]{inputenc} 
\usepackage[T1]{fontenc}    
\usepackage{hyperref}       
\usepackage{url}            
\usepackage{booktabs}       
\usepackage{amsfonts}       
\usepackage{nicefrac}       
\usepackage{microtype}      
\usepackage{pifont}
\usepackage{subcaption}
\usepackage{lipsum}		
\usepackage{graphicx}
\usepackage{natbib}
\usepackage{doi}
\usepackage{float}

\usepackage{listings}
\usepackage{xcolor}
\usepackage{wrapfig}
\usepackage{amsmath}
\renewcommand{\subsubsubsection}[1]{\paragraph{#1}\mbox{}\\}
\definecolor{codegreen}{rgb}{0,0.6,0}
\definecolor{codegray}{rgb}{0.5,0.5,0.5}
\definecolor{codepurple}{rgb}{0.58,0,0.82}
\definecolor{backcolour}{rgb}{0.95,0.95,0.92}
\lstdefinestyle{pythonStyle}{
    language=Python,
    basicstyle=\ttfamily\small,
    keywordstyle=\color{blue},
    stringstyle=\color{codegreen},
    commentstyle=\color{codegray}\itshape,
    numbers=left,
    numberstyle=\tiny\color{codegray},
    numbersep=5pt,
    backgroundcolor=\color{white},
    showspaces=false,
    showstringspaces=false,
    showtabs=false,
    tabsize=4,
    captionpos=b,
    breaklines=true,
    breakatwhitespace=true,
    breakautoindent=true,
    linewidth=\textwidth,
    frame = single,
    backgroundcolor=\color{backcolour}, 
}

\title{\texttt{landmarker}: a Toolkit for Anatomical Landmark Localization in 2D/3D Images}

\author{ \href{https://orcid.org/0000-0003-3608-0308}{\includegraphics[scale=0.06]{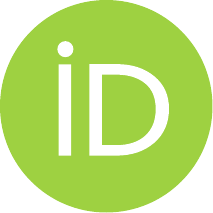}\hspace{1mm}Jef Jonkers}\\
	IDLab \\
        Department of Electronics and Information Systems \\
        Ghent University, Belgium \\
	\texttt{jef.jonkers@ugent.be} \\
	\And
    \href{https://orcid.org/0000-0003-0462-3638}{\includegraphics[scale=0.06]{orcid.pdf}\hspace{1mm}Luc Duchateau}\\
	Biometrics Research Group \\ 
        Department of Morphology, Imaging, Orthopedics, \\
        Rehabilitation and Nutrition \\
        Ghent University, Belgium \\
        \texttt{luc.duchateau@ugent.be} \\
	\And
    \href{https://orcid.org/0000-0001-9530-3466}{\includegraphics[scale=0.06]{orcid.pdf}\hspace{1mm}Glenn Van Wallendael}\\
	IDLab \\
        Department of Electronics and Information Systems \\
        Ghent University - imec, Belgium \\
	\texttt{glenn.vanwallendael@ugent.be} \\
	\And
	\href{https://orcid.org/0000-0002-7865-6793}{\includegraphics[scale=0.06]{orcid.pdf}\hspace{1mm}Sofie Van Hoecke}\\
	IDLab \\
        Department of Electronics and Information Systems \\
        Ghent University - imec, Belgium \\
	\texttt{sofie.vanhoecke@ugent.be} \\
}

\renewcommand{\shorttitle}{\texttt{landmarker}: a Toolkit for Anatomical Landmark Localization in 2D/3D Images}

\hypersetup{
pdftitle={landmarker: a Toolkit for Anatomical Landmark Localization in 2D/3D Images},
pdfsubject={cs.CV, cs.AI, cs.LG},
pdfauthor={Jef Jonkers, Luc Duchateau, Glenn Van Wallendael, Sofie Van Hoecke},
pdfkeywords={Landmark localization, Keypoint detection, Medical image analysis, Python, PyTorch},
}

\begin{document}
\maketitle

\begin{abstract}
	Anatomical landmark localization in 2D/3D images is a critical task in medical imaging. Although many general-purpose tools exist for landmark localization in classical computer vision tasks, such as pose estimation, they lack the specialized features and modularity necessary for anatomical landmark localization applications in the medical domain. Therefore, we introduce \texttt{landmarker}, a Python package built on PyTorch. The package provides a comprehensive, flexible toolkit for developing and evaluating landmark localization algorithms, supporting a range of methodologies, including static and adaptive heatmap regression. \texttt{landmarker} enhances the accuracy of landmark identification, streamlines research and development processes, and supports various image formats and preprocessing pipelines. Its modular design allows users to customize and extend the toolkit for specific datasets and applications, accelerating innovation in medical imaging. \texttt{landmarker} addresses a critical need for precision and customization in landmark localization tasks not adequately met by existing general-purpose pose estimation tools.
    
    \textbf{Code}: \url{https://github.com/predict-idlab/landmarker}.
    
    \textbf{Published\footnote{Published in SoftwareX}:} \url{https://doi.org/10.1016/j.softx.2025.102165}
\end{abstract}

\section{Motivation and significance}

Landmark (or keypoint) localization in 2D/3D images is a fundamental challenge in computer vision, crucial for tasks such as pose estimation~\cite{luvizon_2d3d_2018, tompson_joint_2014, newell_stacked_2016}, face alignment~\cite{zhou_star_2023, kumar_uglli_2019}, robotic manipulation~\cite{ziegler_fashion_2020}, surgical procedures~\cite{sharan_point_2021}, image registration~\citep{grewal_automatic_2023}, various image-based medical diagnoses~\cite{de_queiroz_tavares_borges_mesquita_artificial_2023, ryu_automated_2022, pei_learning-based_2023, payer_integrating_2019, thaler_modeling_2021} and treatment response modeling in longitudinal images~\citep{ou_deformable_2015, cai_deep_2021}. The effectiveness of different approaches is often domain and task-dependent, necessitating extensive experimentation for new applications. Popular methods, such as heatmap regression, involve numerous preprocessing and post-processing steps. The latter, such as sub-pixel accuracy methods, can significantly improve the precision needed in medical applications but nevertheless are frequently overlooked.

As existing packages such as OpenPose~\cite{cao_openpose_2021} and Ultralytics~\cite{jocher_ultralytics_2023} are primarily designed for pose estimation, they lack the modularity needed for medical applications. While MMPose~\cite{mmpose_contributors_openmmlab_2020} offers different components for various tasks, it is still tailored toward pose estimation and does not easily integrate with other frameworks or handle specific medical imaging requirements. Although MMPose \citep{mmpose_contributors_openmmlab_2020} supports 3D landmark localization, it cannot process 3D input data, a critical requirement for medical imaging applications. Furthermore, while MMPose offers some flexibility in model creation, its pipeline mandates the use of its framework for preprocessing, training, and evaluation, limiting compatibility with deep learning frameworks such as PyTorch Lightning~\citep{Falcon_PyTorch_Lightning_2019}. Additionally, neither pose estimation library supports medical imaging formats (e.g., NIfTI and DICOM) or provides specialized data augmentation techniques to simulate common acquisition artifacts. Table~\ref{tab:comparison} summarizes the key differences between these existing packages and the proposed landmark detection package within the medical imaging context.

\newcommand{\cmark}{\textcolor{green!60!black}{\ding{51}}} 
\newcommand{\xmark}{\textcolor{red!70!black}{\ding{55}}}   

\begin{table}[htbp]
\centering
\begin{tabular}{@{}lcccccc@{}}
\toprule
 &  & \multicolumn{2}{c}{3D} &  & \multicolumn{2}{c}{Medical} \\ \cmidrule(lr){3-4}
 Package      & Custom       & Keypoints & Images & Modular & Format & Aug.     \\ 
\midrule
ultralytics \citep{jocher_ultralytics_2023} & \xmark & \xmark & \xmark & \xmark & \xmark & \xmark \\
MMPose \citep{mmpose_contributors_openmmlab_2020}     & \cmark & \cmark & \xmark & \xmark & \xmark & \xmark \\
\textbf{\texttt{landmarker}}  & \cmark & \cmark & \cmark & \cmark & \cmark & \cmark \\
\bottomrule
\end{tabular}%
\caption{Comparison of Python packages for keypoint/landmark localization.}
\label{tab:comparison}
\end{table}

Our Python package, \texttt{landmarker}, addresses the need for precise anatomical landmark localization in medical images, which is critical for diagnostics and therapeutic procedures in different specialties, such as orthodontics, maxillofacial surgery, and orthopedics. It is a flexible PyTorch-based toolkit designed explicitly for training, modeling, and evaluating anatomical landmark localization problems. This toolkit accelerates the development of algorithms and enhances the accuracy of landmark identification in medical images, thereby improving diagnostic and treatment outcomes. Its modular and adaptable framework allows researchers and practitioners to implement state-of-the-art algorithms tailored to their specific datasets.

\texttt{landmarker} has a modular framework, which supports various landmark localization algorithms. This flexibility enables customization and extension according to specific needs. The package offers interfaces for data preprocessing, model training, evaluation, and visualization.

\section{Background: landmark localization}
\begin{wrapfigure}{r}{0.5\textwidth}
    \centering
    \includegraphics[width=0.5\textwidth]{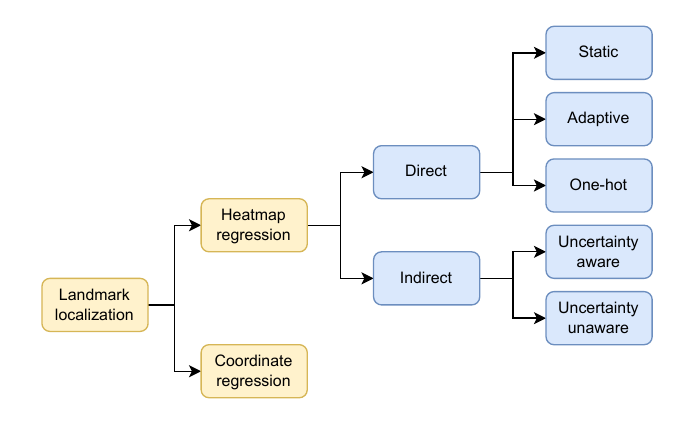}
    \caption{Taxonomy of (deep learning) landmark localization approaches. The frequently used taxonomy in the literature is highlighted in yellow, while our extended taxonomy of the problem is highlighted in blue.}
    \label{fig:types}
\end{wrapfigure}

There are two primary methodologies in landmark localization (see Figure \ref{fig:types}): heatmap regression and coordinate regression. Coordinate regression~\citep{fard_acr_2022} involves learning a direct relationship between the image and the landmark locations. However, the introduction of heatmap regression~\cite{tompson_joint_2014}, which generally yields better performance, has shifted the research focus. Heatmap regression
predicts a heatmap that assigns a likelihood or probability to each (downsized) image pixel. Early heatmap regression methods, termed \textit{direct} heatmap regression, assume a parametric distribution, typically a bivariate Gaussian or Laplacian distribution. Direct heatmap regression can be further classified into static and adaptive approaches. In static heatmap regression, the heatmap distribution parameters are fixed hyperparameters during training. In adaptive heatmap regression, these parameters are adjusted during training, either by treating them as learnable model parameters~\citep{payer_integrating_2019, payer_uncertainty_2020, thaler_modeling_2021} or by using scheduling methods that modify the parameters based on specific evaluation metrics~\citep{teixeira_adaloss_2019}. Another variant of static heatmap regression involves training a model on one-hot heatmap images (masks), essentially parameterizing the heatmap as a categorical distribution and transforming the task into a multi-class classification problem where each pixel represents a class~\citep{mccouat_contour-hugging_2022}.

Recent advancements on heatmap regression have seen the rise of fully convolutional neural networks (CNNs) with differentiable decoding operations, such as the soft-argmax operation~\cite{luvizon_2d3d_2018}. These methods, prominent in facial landmark localization, use the heatmap generation layer as an intermediate layer and the decoding operation as the final layer, optimizing the network with a loss function that compares the decoded coordinates to the ground-truth coordinates. Some of these methods account for the ambiguity in landmark ground truth by incorporating uncertainty-aware loss functions, such as the Gaussian log-likelihood loss~\citep{kumar_uglli_2019}, or other custom loss functions~\cite{kumar_uglli_2019, kumar_luvli_2020, zhou_star_2023}.

In the medical field, landmark localization often utilizes two-stage approaches \citep{jiang_cephalformer_2022, song_automatic_2020, zhong_attention-guided_2019,nguyen_intelligent_2020}. The first stage identifies landmarks on a low-resolution image, while the second stage refines this localization using a high-resolution patch, which is a region of interest based on the initial predictions. One of the previously described approaches can be applied in both stages. Another technique, also used in medical imaging, involves segmenting or contouring specific anatomical structures and inferring the landmarks from the resulting masks or contours \citep{goosen_model-based_2011, pei_automated_2021}.

The diversity of approaches in landmark localization, from basic coordinate regression to sophisticated uncertainty-aware heatmap methods, highlights the complexity of this field, particularly in medical imaging applications. While general-purpose tools like OpenPose and MMPose exist, they primarily focus on human pose estimation and lack the specialized features needed for medical imaging applications. These tools often do not support the full range of heatmap generation and decoding methods, uncertainty quantification, or medical image-specific preprocessing pipelines crucial for anatomical landmark localization. Additionally, existing tools typically do not provide the modularity required to implement and experiment with different approaches, such as two-stage refinement methods commonly used in medical applications. These limitations, combined with the need for high precision in medical contexts, underscore the need for a specialized toolkit to handle the unique challenges of anatomical landmark localization while supporting established and emerging methodologies.

\section{Software description}
\texttt{landmarker} is a Python package that leverages PyTorch \cite{paszke_pytorch_2019} deep learning framework and MONAI \cite{cardoso_monai_2022}, a PyTorch-based deep learning framework for healthcare imaging for handling medical image files and transformations.

Users can install \texttt{landmarker} via \href{https://pypi.org/project/landmarker/}{pip}. After installation, our \href{https://predict-idlab.github.io/landmarker/}{documentation} and variety of \href{https://github.com/predict-idlab/landmarker/tree/main/examples}{examples} will guide users in applying the toolkit for their specific needs.

\subsection{Software architecture and functionalities}
\begin{figure}
    \centering
    \includegraphics[width=\textwidth]{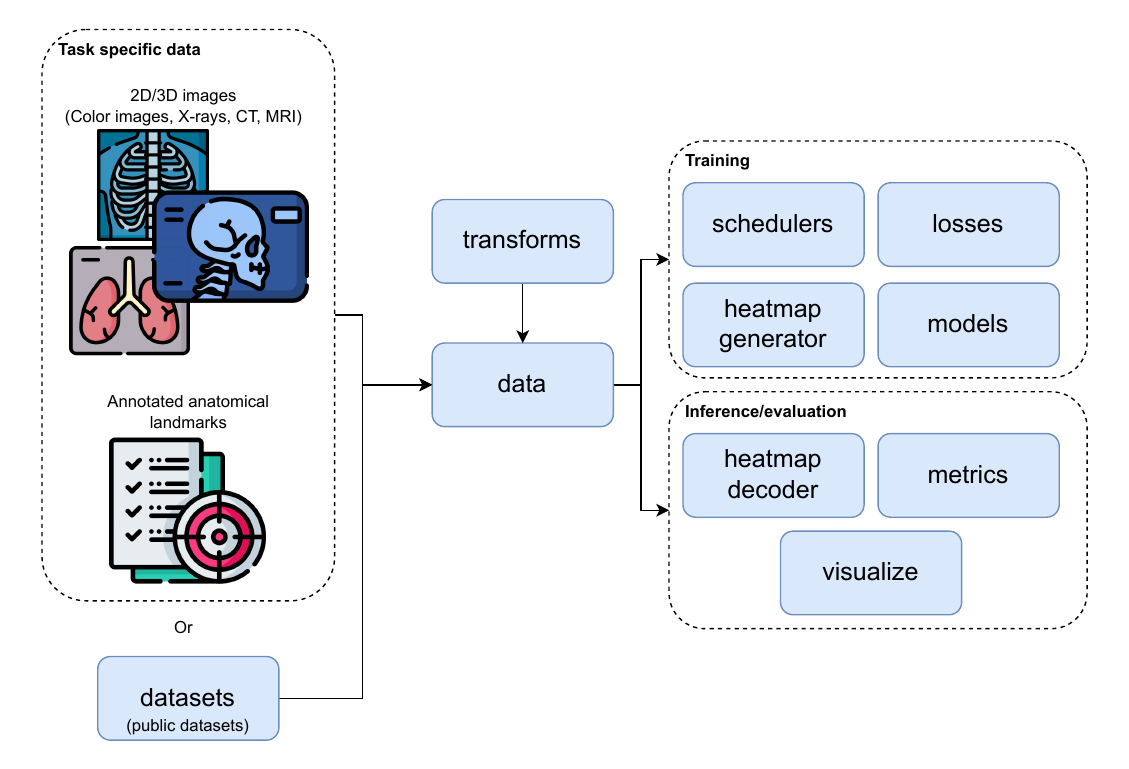}
    \caption{Flowchart of the different sub-packages of the \texttt{landmarker} pacakge.}
    \label{fig:flowchart}
\end{figure}

The \texttt{landmarker} package is structured into several key modules, each addressing specific aspects of the landmark localization pipeline, see Figure~\ref{fig:flowchart}.

\subsubsection{Data handling and preprocessing}
The \texttt{data} module provides flexible \textit{"dataset"} classes that inherit from PyTorch's \texttt{Dataset}, ensuring compatibility with PyTorch \texttt{DataLoaders}. Four main dataset types are supported:
\begin{itemize}
    \item \texttt{LandmarkDataset}: For images with corresponding landmark coordinates.
    \item \texttt{HeatmapDataset}: For images with corresponding static heatmaps representing landmarks.
    \item \texttt{MaskDataset}:  For images with binary segmentation masks indicating landmark locations.
    \item \texttt{PatchDataset}: For image patches or regions of interest (ROI) with corresponding landmarks.
\end{itemize}
These datasets support various image formats (NIfTI, DICOM, PNG, JPG, BMP, NPY/NPZ) through integration with the MONAI library \citep{cardoso_monai_2022}. The module can handle single-class, multi-class single-instance, and multi-class multi-instance landmark problems. Utility functions are provided to transform common annotation formats (e.g., labelme~\cite{wada_labelme_2024}) into the required format.

The package also includes the \texttt{datasets} module, which contains functions to import benchmark datasets, such as the ISBI2015~\cite{wang_benchmark_2016} dataset directly. Preprocessing and data augmentation capabilities are available, leveraging MONAI's transformations for 2D and 3D data. The package handles affine landmarks transformations through the \texttt{transforms} module.

\subsubsection{Heatmap generation and decoding}
The \texttt{heatmap} module provides functionality for generating target heatmaps and decoding predicted heatmaps, which is essential for heatmap regression approaches.

\subsubsubsection{Heatmap generation}
The \texttt{HeatmapGenerator} class and its subclasses support various parametric distributions (e.g., multivariate Gaussian and Laplace). The design allows easy implementation of custom distributions. The parameters of the parametric distribution can be set on initialization. Still, they can also be changed during training, for example, for adaptive heatmap regression approaches that rely on a scheduler, such as Adaloss scheduler \citep{teixeira_adaloss_2019}. Additionally, the parameters could also be learnable parameters during a training procedure.

\subsubsubsection{Heatmap decoding}
While heatmap generation is only needed for direct approaches, decoding heatmaps is needed for all heatmap regression approaches, with the caveat that the indirect approaches need differentiable decoding operations. Multiple decoding methods are implemented: 
\\

The \textit{argmax} operation is the simplest decoder operation~\cite{tompson_joint_2014}. The operation takes the maximum value of the heatmap $H_i$,
\begin{equation}
    \hat{y}_i = \text{argmax} H_i,
\end{equation}
to get the coordinate $\hat{y}_i$ of landmark $i$.
The major downside of this approach is that it introduces a discretization error by only choosing a pixel as the predicted landmark.
\\

The \textit{weighted spatial mean} \citep{luvizon_2d3d_2018, dong_supervision-by-registration_2018, kumar_uglli_2019, kumar_luvli_2020} is the approach for indirect heatmap generation. The operation calculates the landmark location $\hat{y}_i$ by first post-processing the heatmap $H_i$ with an activation function $\sigma$, typically a type of normalization operation, and afterward taking the weighted spatial mean, e.g., the two-dimensional case:
\begin{equation}
    \hat{y}_i = \left(\sum_{c=0}^W\sum_{l=0}^H \frac{c}{W} \sigma(H_i)_{l,c}, \sum_{c=0}^W \sum_{l=0}^H \frac{l}{H} \sigma(H_i)_{l,c}\right)
\end{equation}
where $W$ and $H$ are, respectively, the heatmap's width and height.
Frequently used activation functions $\sigma$ are a ReLU activation combined with a normalization procedure and the softmax operation on the heatmap, which is in the literature often referred to as soft-argmax operation \citep{luvizon_2d3d_2018, dong_supervision-by-registration_2018, bulat_subpixel_2021}.
\\

A significant issue of taking the weighted spatial mean is that it can globally lead to semantically unstructured outputs and thus decrease performance \citep{bulat_subpixel_2021}. In \cite{bulat_subpixel_2021}, they propose to apply soft-argmax operation locally, i.e., a small window, around the heatmap location with a maximum heatmap value. We implement this method and extend it to be used with any post-processing function, i.e., \textit{local weight spatial mean} operation.

\subsubsection{Models and losses}
While \texttt{landmarker} is compatible with any PyTorch model or loss function, it includes implementations of successful approaches from the literature, such as the spatial configuration network \cite{payer_integrating_2019}, which is effective for positionally consistent medical images like cephalograms. Also, multiple heatmap regression loss functions are implemented from successful approaches \citep{thaler_modeling_2021, nibali_numerical_2018, kumar_uglli_2019, feng_wing_2018, wang_adaptive_2019, zhou_star_2023}.

\subsubsection{Evaluation and visualization}
The \texttt{visualize} module allows inspecting constructed datasets and trained models. Additionally, it enables the generation of detection reports that output several metrics from the \texttt{metrics} module, such as point error and success detection rate (SDR), which is the percentage of predicted landmarks with a point error smaller or equal to a specified radius.

\section{Illustrative examples}
In this section, we illustrate the use of \texttt{landmarker}. We will perform adaptive heatmap regression with the ISBI2015 cephalometric landmark dataset, showcasing most functionality.

We start by loading the data into a \texttt{LandmarkDataset} by providing a list of paths to images, a NumPy array of the shape $(N,C,D)$ where $N$ is the number of samples, $C$ is the number of landmark classes, and $D$ represents the spatial dimension which can be 2 or 3.

\begin{lstlisting}[language=Python, breaklines=true, caption=Loading data into a \texttt{LandmarkDataset}]
from landmarker.data import LandmarkDataset

image_paths = ... # list of paths of images
landmarks = ... # numpy array or torch tensor provided
pixel_spacing = ... # pixel spacing of the images
class_names = ... # names of the landmarks
compose_transform = ... # MONAI compose transform

ds = LandmarkDataset(
    imgs=image_paths,
    landmarks=landmarks,
    spatial_dims=2,
    transform=compose_transform,
    dim_img=(512,512) # dimension to resize to
    class_names=class_names)
\end{lstlisting}

Alternatively, one can also use the \texttt{dataset} module, which includes everything to import the ISBI2015 dataset directly.

\begin{lstlisting}[language=Python, breaklines=true, caption=Loading data into a \texttt{LandmarkDataset} through the \texttt{dataset} module.]
from landmarker.dataset import get_cepha_landmark_datasets

data_dir = ...
ds_train, ds_test1, ds_test2 = get_cepha_landmark_datasets(data_dir)
\end{lstlisting}

Once the data is loaded in the proper format, we can set up the heatmap generator to generate a heatmap from landmarks.

\begin{lstlisting}[language=Python, breaklines=true, caption=Setting up Gaussian heatmap generator.]
from landmarker.heatmap import GaussianHeatmapGenerator

generator = GaussianHeatmapGenerator(
    nb_landmarks=19, # number of landmarks
    sigmas=3, # sd. value of the Gaussian heatmap function
    learnable=True, # make the covariance matrix learnable parameters
    heatmap_size=(512,512))
\end{lstlisting}

\begin{figure}
    \centering
    \includegraphics[width=\textwidth]{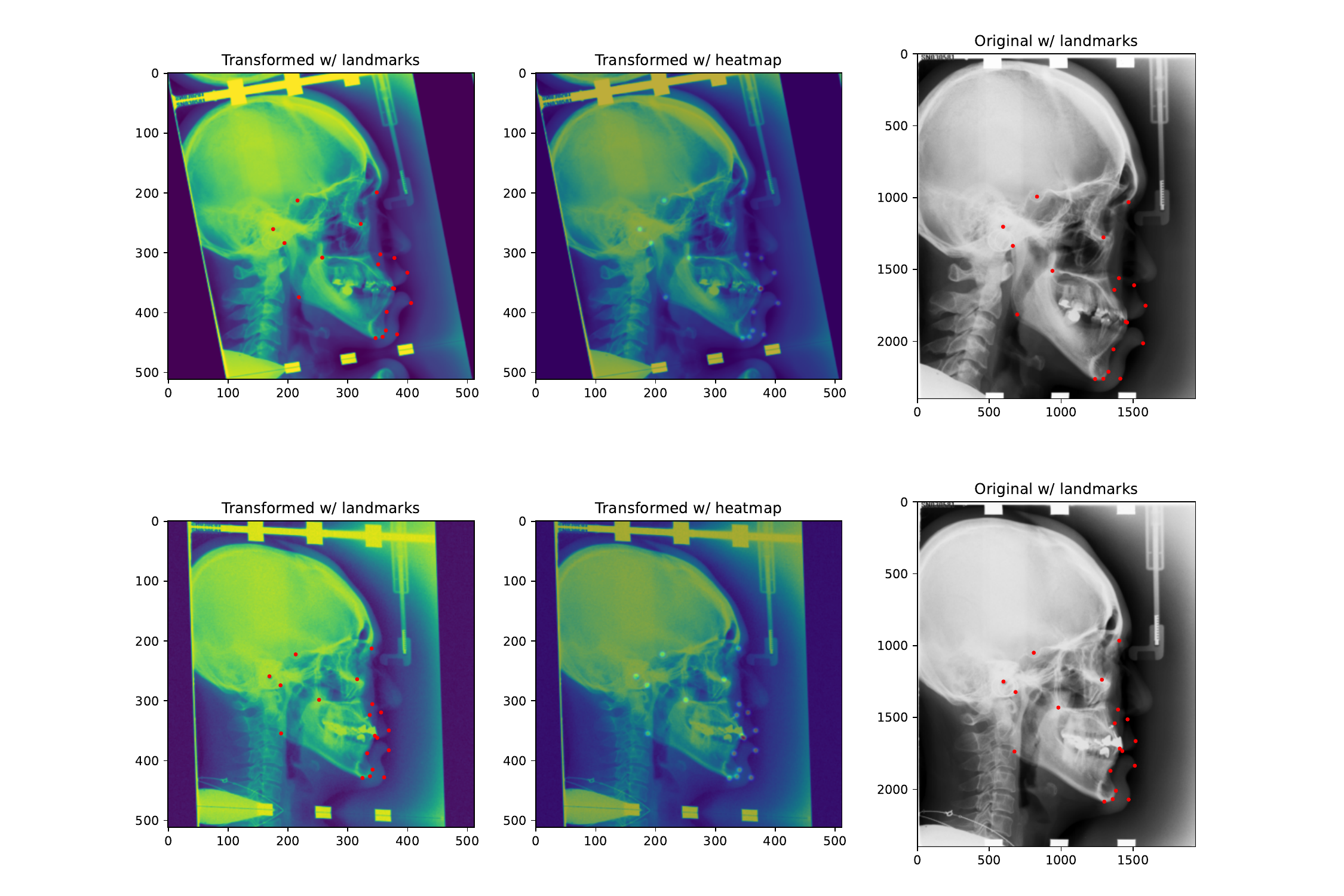}
    \caption{Example of results of running the \texttt{inspection\_plot} on the ISBI2015 dataset.}
    \label{fig:inspect}
\end{figure}

After this, the user can visually inspect if everything is initialized as expected by using the \texttt{inspection\_plot} function (Listing \ref{lst:inspect} and Figure \ref{fig:inspect}).
\begin{lstlisting}[language=Python, breaklines=true, caption=Loading data into a \texttt{LandmarkDataset}, label=lst:inspect]
from landmarker.visualize import inspection_plot

inspection_plot(ds_train, range(3), heatmap_generator=heatmap_generator)
\end{lstlisting}

With the \texttt{LandmarkDataset} and \texttt{HeatmapGenerator} adequately set up, we can now train a model to output likelihood heatmaps. In Listing \ref{lst:training}, we train a SpatialConfigurationNetwork~\cite{payer_integrating_2019, thaler_modeling_2021} and parameterization of the Gaussian heatmap~\cite{payer_uncertainty_2020, thaler_modeling_2021}.

\begin{lstlisting}[language=Python, breaklines=true, caption=Training adaptive heatmap regression model., label=lst:training]
import torch

from landmarker.models import OriginalSpatialConfigurationNet
from landmarker.losses import GaussianHeatmapL2Loss

model = OriginalSpatialConfigurationNet(in_channels=1, out_channels=19)
optimizer = torch.optim.SGD([
{'params': model.parameters(), "weight_decay":1e-3},
{'params': heatmap_generator.sigmas},
{'params': heatmap_generator.rotation}]
, lr=1e-6, momentum=0.99, nesterov=True)

criterion = GaussianHeatmapL2Loss(alpha=5)

train_loader = DataLoader(ds_train, batch_size=1, shuffle=True, num_workers=0)

# Start training
for epoch in range(100): 
    model.train()
    for batch in train_loader:
        images = batch["image"].to(device)
        landmarks = batch["landmark"].to(device)
        optimizer.zero_grad()
        outputs = model(images)
        heatmaps = heatmap_generator(landmarks)
        loss = criterion(outputs, heatmap_generator.sigmas, heatmaps)
        loss.backward()
        optimizer.step()
\end{lstlisting}

\begin{figure}[htbp]
    \centering
    \begin{subfigure}[b]{\textwidth}
         \includegraphics[width=\textwidth]{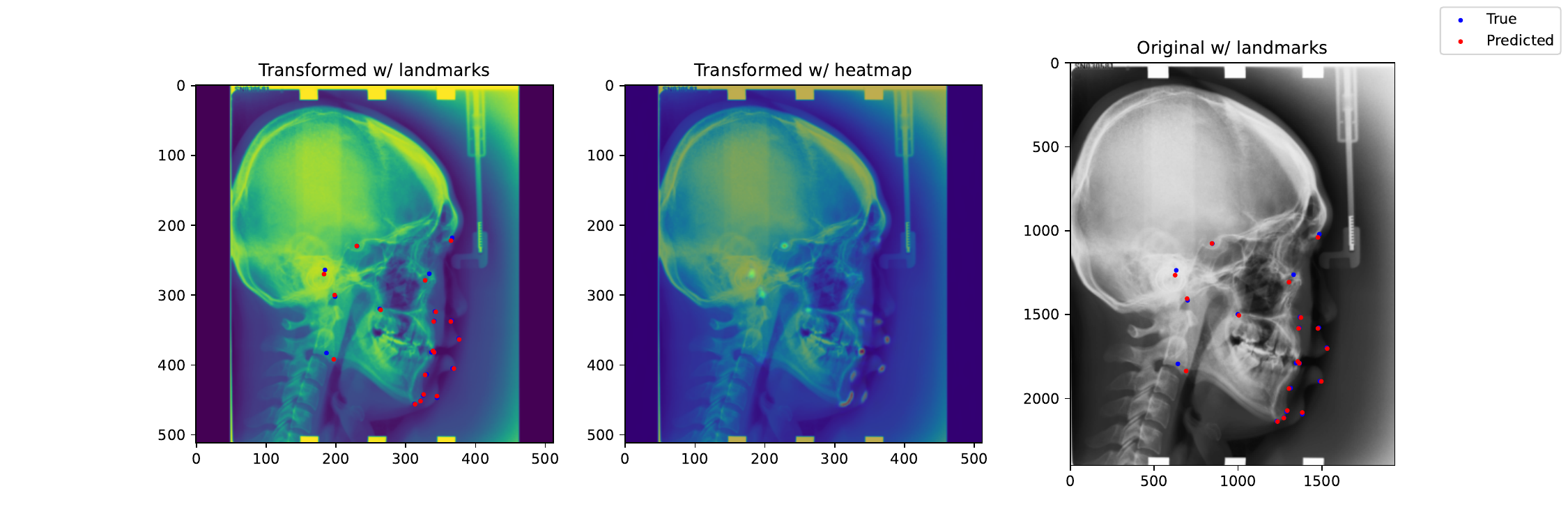}
         \caption{ISBI2015 dataset}
         \label{fig:pred_inspect_isbi2015}
    \end{subfigure}
    \vspace{1em} 
    \begin{subfigure}[b]{\textwidth}
         \includegraphics[width=\textwidth]{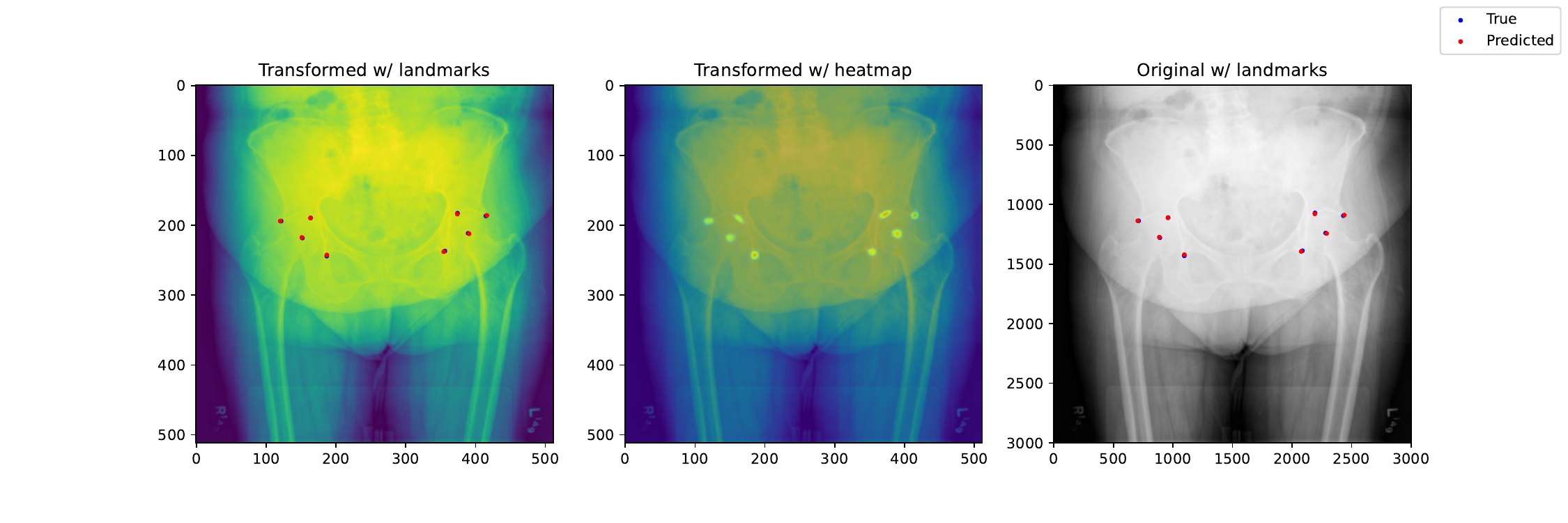}
         \caption{Pelvis X-rays}
         \label{fig:pred_inspect_pelvis}
    \end{subfigure}
    \begin{subfigure}[b]{\textwidth}
         \includegraphics[width=\textwidth]{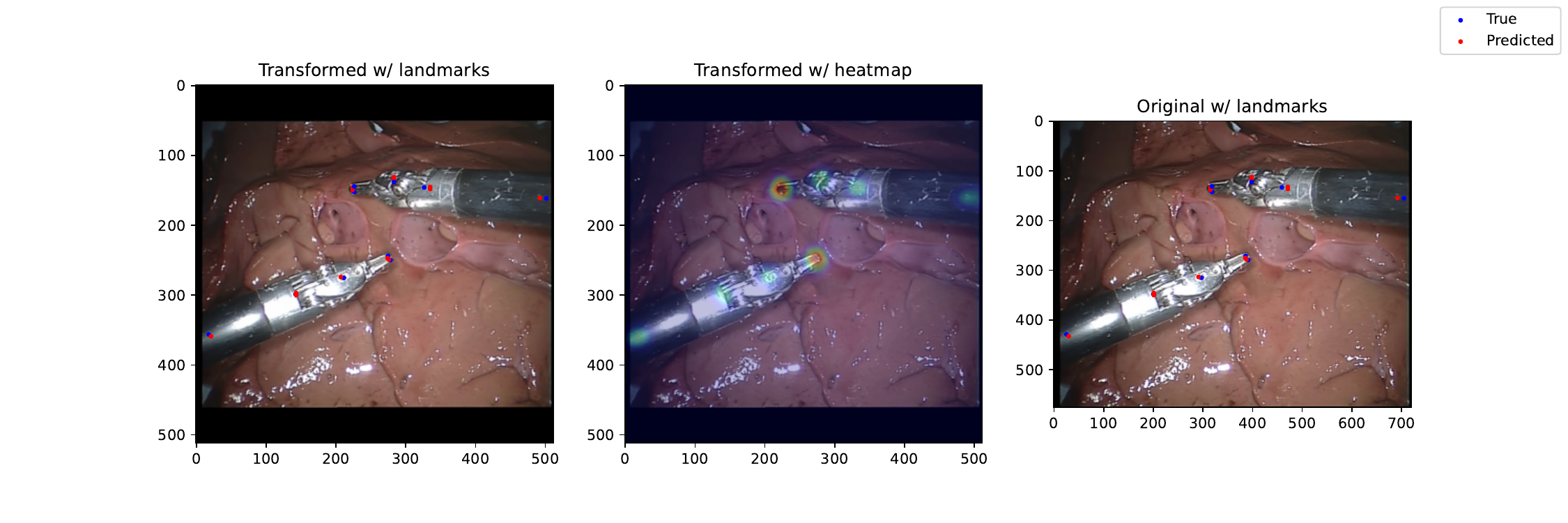}
         \caption{Endoscopic images}
         \label{fig:pred_inspect_endovis}
    \end{subfigure}
    \vspace{1em} 
    \caption{Examples of landmark predictions using the \texttt{prediction\_inspection\_plot} tool for different datasets: (a) Endoscopic images, (b) ISBI2015, and (c) Pelvis X-rays.}
    \label{fig:pred_inspect}
\end{figure}

After the training procedure, one can visually evaluate the results by using the \texttt{prediction\_inspect\_plot} function (see Listing \ref{lst:pred_inspect} and Figure \ref{fig:pred_inspect}) or by creating a detection report.

\begin{lstlisting}[language=Python, breaklines=true, caption=Prediction inspection plot., label=lst:pred_inspect]
from landmarker.visualize import prediction_inspect_plot

prediction_inspect_plot(ds_test1, model, ds_test1.indices[:3])
\end{lstlisting}

\subsection{Benchmarks}
To evaluate our package capabilities and its model's landmark localization precision quantitatively, we conducted two benchmark experiments on relatively recent datasets that showcase its performance on both 2D and 3D medical imaging tasks. For the 2D benchmark, we used pelvis x-ray images from the Osteoarthritis Initiative \citep{pei_learning-based_2023}. In this experiment, our spatial configuration network (SCN) with adaptive heatmaps—implemented in \texttt{landmarker}, was compared against a state-of-the-art U-Net with Attention approach \citep{pei_learning-based_2023}. As shown in Table \ref{tab:pelvis-sota}, our method achieved a lower average position error (PE) and higher success detection rates (SDR) across multiple thresholds, clearly demonstrating the capabilities of the package as a benchmarking tool.

\begin{table}[htp]
\centering
\caption{Landmark localization results of spatial configuration network (SCN) with adaptive heatmaps \citep{payer_uncertainty_2020, thaler_modeling_2021} (implemented in \texttt{landmarker}) on the pelvis x-rays (2D) of the osteoarthritis initiative dataset \citep{pei_learning-based_2023} compared against the literature. The best results are highlighted in bold. A 5-fold validation approach, as suggested by Pei et al. \citep{pei_learning-based_2023}, obtains the results.}
\label{tab:pelvis-sota}
\resizebox{0.6\textwidth}{!}{
\begin{tabular}{llllll}
\hline
Model name        & PE (mm) & \multicolumn{4}{c}{SDR (\%)}                  \\ \cline{3-6} 
                  &         & 1 mm    & 2 mm  & 3 mm    & 4 mm    \\ \hline
\begin{tabular}[c]{@{}l@{}}U-Net w/ Attention \citep{pei_learning-based_2023}\end{tabular} & 3.14    & 8.65\% & 33.89\% & 59.25\% & 78.61\% \\ \hline
\begin{tabular}[c]{@{}l@{}}SCN \citep{payer_uncertainty_2020, thaler_modeling_2021} \\ (\texttt{landmarker}) \end{tabular} &
  \textbf{1.61} &
  \textbf{33.25\%} &
  \textbf{74.38\%} &
  \textbf{90.96\%} &
  \textbf{95.85\%} \\ \hline
\end{tabular}
}
\end{table}

In addition to the 2D evaluation, we benchmarked our model on 3D images using CT scans of the skull \citep{he_anchor_2024}, where the task was to annotate mandibular landmarks automatically. The results, presented in Table \ref{tab:mml-sota}, include both validation and test metrics, with our one-hot ensemble \citep{jonkers_reliable_2025} method outperforming the competitive baseline regarding PE and SDR. These results confirm our model's precision and highlight its flexibility in handling diverse anatomical landmarking applications across different imaging modalities.
\begin{table}[htp]
\centering
\caption{Landmark localization results of one hot ensemble \citep{jonkers_reliable_2025} (implemented in \texttt{landmarker}) on the MML (3D) dataset compared against the literature. This benchmark dataset is the same as the data
subset, only with complete landmarks, used as a benchmark in He et al. \citep{he_anchor_2024}. The best results are highlighted in bold.}
\label{tab:mml-sota}
\resizebox{\textwidth}{!}{
\begin{tabular}{llllll|lllll}
\hline
Model name        & \multicolumn{5}{c|}{Validation}                          & \multicolumn{5}{c}{Test}                        \\ \cline{2-11} 
                  & PE (mm) & \multicolumn{4}{c|}{SDR (\%)}                  & PE (mm) & \multicolumn{4}{c}{SDR (\%)}          \\ \cline{3-6} \cline{8-11} 
                  &         & 2 mm    & 2.5 mm  & 3 mm    & 4 mm             &     & 2 mm    & 2.5 mm  & 3 mm    & 4 mm    \\ \hline
\begin{tabular}[c]{@{}l@{}}Pruning-ResUNet3D \citep{he_anchor_2024}\end{tabular} & 1.82    & 73.21\% & 82.14\% & 88.93\% & \textbf{94.76\%} & 1.96    & 70.03\% & 79.97\% & 86.10\% & 92.73\% \\ \hline
\begin{tabular}[c]{@{}l@{}}\textbf{One hot ensemble \citep{jonkers_reliable_2025}} \\ (\texttt{landmarker}) \end{tabular} &
  \textbf{1.60} &
  \textbf{77.81\%} &
  \textbf{87.76\%} &
  \textbf{89.92\%} &
  93.37\% &
  \textbf{1.39} &
  \textbf{81.67\%} &
  \textbf{91.31\%} &
  \textbf{93.33\%} &
  \textbf{96.31\%} \\ \hline
\end{tabular}
}
\end{table}

\section{Impact and conclusions}
The development and release of the \texttt{landmarker} Python package offer significant advancements in anatomical landmark localization, particularly within medical imaging. By providing a flexible and modular toolkit tailored explicitly for medical imaging applications, \texttt{landmarker} addresses a critical need for precision and customization in landmark localization tasks that are not adequately met by existing general-purpose pose estimation tools like OpenPose or MMPose.

One of the key impacts of \texttt{landmarker} is its ability to streamline the research and development process for researchers and developers working on image-based diagnostic and therapeutic deep learning applications. The package's integration with PyTorch and MONAI ensures compatibility with various medical image formats and processing pipelines, facilitating easy adoption in research environments. Moreover, the extensive support for different landmark localization methodologies, including static and adaptive heatmap regression, allows users to implement state-of-the-art algorithms with minimal overhead.

Including customizable modules for data handling, preprocessing, heatmap generation, and model evaluation further enhances the utility of \texttt{landmarker}. These features enable users to quickly adapt the package to their specific use cases, whether they involve 2D or 3D images, single-class or multi-class landmark localization, or static versus adaptive methodologies. The ability to implement and test novel approaches within a unified framework accelerates innovation in computer-aided diagnosis through medical imaging, potentially leading to more accurate diagnoses and better patient outcomes.

In conclusion, \texttt{landmarker} represents a powerful tool for advancing the field of anatomical landmark localization in medical imaging. Its flexibility and ease of use make it a helpful resource for researchers and engineers. We plan to incorporate future developments and improvements, particularly in uncertainty quantification, which is critical for enhancing the reliability and robustness of landmark localization in medical applications.

\section*{CRedit Authorship Contribution Statement}
\textbf{Jef Jonkers:} Conceptualization, Methodology, Software, Validation, Visualization, Writing - Original Draft. \textbf{Luc Duchateau:} Conceptualization, Writing - Review \& Editing, Supervision, Funding acquisition. \textbf{Glenn Van Wallendael:} Conceptualization, Writing - Review \& Editing, Supervision. \textbf{Sofie Van Hoecke} Conceptualization, Writing - Review \& Editing, Supervision, Funding acquisition.

\section*{Funding sources}
Jef Jonkers is funded by the Research Foundation Flanders (FWO, Ref. 1S11525N). This research was (partially) funded by the Flemish Government (AI Research Program).



\clearpage
\bibliographystyle{unsrtnat} 
\bibliography{references}

\end{document}